\documentclass[letterpaper, 10 pt, conference]{ieeeconf}  

\IEEEoverridecommandlockouts                              

\usepackage{xcolor}
\usepackage{graphicx}
\usepackage{subfig}
\usepackage{amsmath}
\usepackage{amsthm} 
\usepackage{amssymb}
\usepackage{url}

\usepackage{breakurl}
\usepackage[font=footnotesize]{caption} 
\usepackage{subfig} 
\usepackage[noadjust]{cite} 
\usepackage{algorithmic}
\usepackage[linesnumbered,ruled,vlined]{algorithm2e}
\usepackage{enumerate}
\usepackage{threeparttable}
\usepackage{booktabs}
\usepackage[hidelinks,colorlinks=false]{hyperref}
\usepackage[left=0.75in, right=0.75in, top=0.75in, bottom=0.75in]{geometry}
\usepackage{authblk} 
\usepackage{arydshln} 

\usepackage{hyperref}

\hypersetup{
    colorlinks=true,
    linkcolor=blue,
    filecolor=magenta,      
    urlcolor=blue,
    pdftitle={Your Paper Title},
    pdfpagemode=FullScreen,
}

\usepackage{bm}
\usepackage{tikz}
\usetikzlibrary{calc} 
\usetikzlibrary{shapes} 
\usetikzlibrary{chains}
\usetikzlibrary{fit}
\usetikzlibrary{arrows}
\usetikzlibrary{decorations.text} 
\usetikzlibrary{decorations.markings}
\usetikzlibrary{decorations.pathmorphing} 
\usetikzlibrary{shadows}
\usetikzlibrary{patterns}
\usetikzlibrary{matrix}
\usepackage{pgfplots}
\usepackage[europeanresistors]{circuitikz}
\usepackage[outline]{contour} 
\contourlength{1.5pt}

\graphicspath{{figures/}}

\title{\LARGE \bf
SeeBelow: Sub-dermal 3D Reconstruction of Tumors with Surgical Robotic Palpation and Tactile Exploration
}

\author{ Raghava Uppuluri$^{+}$, Abhinaba Bhattacharjee$^{+}$, Sohel Anwar, and Yu She$^*$
\thanks{*Corresponding Author}
\thanks{$^+$ Primary Authors with equal contributions}
\thanks{Raghava Uppuluri is with the School of Electrical and Computer Engineering, College of Engineering, Purdue University,
        West Lafayette, IN 47906 USA.
        {\tt\small ruppulur@purdue.edu}}%
\thanks{Abhinaba Bhattacharjee is with the School of Mechanical Engineering, Purdue University, West Lafayette, IN 47906 USA.
        {\tt\small bhatta13@purdue.edu }}%
\thanks{Sohel Anwar is with the Department of Mechanical and Energy Engineering, School of Engineering and Technology,Purdue University, Indianapolis, IN 46202 USA.
{\tt\small sanwar@purdue.edu}}%
\thanks{Yu She is with the School of Industrial Engineering, College of Engineering, Purdue University,
        West Lafayette, IN 47906 USA.
        {\tt\small yushe@purdue.edu}}%
}

\begin{document}

\maketitle
\thispagestyle{empty}
\pagestyle{empty}

\begin{abstract}
Surgical scene understanding in Robot-assisted
Minimally Invasive Surgery (RMIS) is highly reliant on visual cues and lacks tactile perception. 
Force-modulated surgical palpation with tactile feedback is necessary for localization, geometry/depth estimation, and dexterous exploration of abnormal stiff inclusions in subsurface tissue layers.
Prior works explored surface-level tissue abnormalities or single layered tissue-tumor embeddings with more than 300 palpations for dense 2D stiffness mapping.
Our approach focuses on 3D
reconstructions of sub-dermal tumor surface profiles in multi-layered tissue (skin-fat-muscle) using a visually-guided novel tactile navigation policy.
A robotic palpation probe with tri-axial force sensing was leveraged for tactile exploration of the phantom.
From a surface mesh of the surgical region initialized from a depth camera, the policy explores a surgeon's region of interest through palpation, sampled from bayesian optimization.
Each palpation includes contour following using a contact-safe impedance controller to trace the
sub-dermal tumor geometry, until the
underlying tumor-tissue boundary is reached. Projections of these contour
following palpation trajectories allows 3D reconstruction of
the subdermal tumor surface profile in less than 100 palpations. Our
approach generates high-fidelity 3D
surface reconstructions of rigid tumor embeddings in tissue layers with isotropic elasticities, although soft tumor geometries are yet to be explored. 
For more details, please refer to our open-source codebase\footnote{\href{https://github.com/raghavauppuluri13/seebelow}{https://github.com/raghavauppuluri13/seebelow}} and project website\footnote{\href{https://raghavauppuluri13.github.io/seebelow.github.io/}{https://raghavauppuluri13.github.io/seebelow.github.io/}}.
\end{abstract}

\section{Introduction}

Surgical palpation is a widely performed examination
technique to assess the location and geometry of tissue anomalies during laparoscopic or minimally invasive surgery (MIS). 
Although prior knowledge about surgical tissue conditions (tumors, cysts, lesions, nodules, etc) are available preoperatively from CT, MRI or Ultrasound scan-based medical imaging modalities, the intraoperative surgical scene understanding in robot-assisted minimally invasive surgery (RMIS) is highly challenging. 
Localization of soft tissue conditions in RMIS is  immensely impacted by variances in tissue compliance, peripheral tissue intrusions along the navigating direction and visual occlusions from body fluids.
These factors assert spatial uncertainties in graphical rendering of anatomical landmarks as compared to referenced medical images. 
Surgical palpation of the target anatomy with kinesthetic feedback from surgeon's fingers, as used in conventional open surgery, better assists in the surgical scene understanding and tissue manipulation which is often lost in RMIS. 
The deficiency of tactile perception in RMIS can be compensated with robotic surgical palpation probes which will add tactile information about the target tissue stiffness and compliance. This information can demarcate boundaries of underlying tissue abnormalities while reconstructing the 3D surface profiles of subdermal stiffer tumors.
 
\section{Related Work}

Previous works investigated different force sensing modalities\cite{5732701,jeong2020miniature,xie2014optical,li2022compact} to study soft compliant surface contact interactions \cite{nagy2019recent, wijayarathne2023real} and force sensing uncertainty compensations \cite{guo2019compensating} for soft tissue palpation and exploration. 
Recent advancements in commercial surgical platforms\cite{miller2019intuitive,leonard2014smart} achieved amazing feats \cite{nifong2003robotic} in execution of semi-autonomous surgical procedures \cite{misal2021robotic}. 

Nonetheless, adequate robotic surgical autonomy might be reached with augmenting the perception of robotic touch with surgical haptics\cite{culmer2020haptics} and force modulated tactile imaging \cite{li2020reaction, liu2009rolling, suresh2022shapemap}. 
A complimentary approach of facilitating high resolution robotic touch with HD in-vivo stereo-vision for laparoscopic robotic surgery will imitate a surgeon’s hand-eye coordination-based dexterity in precise form factor to better navigate and map intricate anatomical structures of the tissue. Autonomous robotic probing of tumor tissues with active area search \cite{salman2018trajectory} have found promising results optimizing over a Gaussian Process (GP) regression model built from discrete palpations measuring stiffness of random indentation points. 
Gaussian Process regression (GP) \cite{garg2016tumor} and Bayesian Optimization (BO) \cite{ayvali2016using} has been effective in reducing global search space minimizing computation cost and palpation time for indentation based probing to locate stiff inclusion in compliant surfaces. 
In RMIS, the robotic laparoscopic probes are kinematically constrained within the  intraoperative space. 
Therefore surgical planning needs safe contact-rich manipulations applying  optimal forces to the  target tissues and their underlying layers which have unmodelled characteristics (e.g varying contact friction, slip, stiffness, and damping). 

Soft tissue scanning is extensively practiced in orthopaedic manual therapy to assess pain in chronic musculoskeletal pain conditions. In context, classified force-motion patterns \cite{bhattacharjee2023multimodal} of quantifiable soft tissue manipulations\cite{bhattacharjee2021quantifiable} using a handheld localized \cite{bhattacharjee2021data} and dispersive force-motion applicators \cite{bhattacharjee2022handheld,bhattacharjee2022finite} are studied for clinical manual therapy treatment assessments. Palpation techniques \cite{robinson2009reliability} adopted from handheld instrumented soft tissue manipulations\cite{kim2017therapeutic,gulick2014influence} can form the basis of dynamic robotic tactile probing for trajectory optimized geometry estimation of sub-dermal surgical tumor localization.

TIn context, our prior work incorporated finite element modeling and simulation of quasi-static robotic palpation \cite{10.1115/1.4063470} to investigate strain and deformations in underlying subsurface tissue layers due to the presence and absence of a stiff adipose sarcoma (tumor). 
Results revealed that presence of a stiffer tumor reduces strain distribution over peripheral tissue layers indicating negligible principal strains in layers underneath the tumor surface as evident from their deformation profiles. 
This study is an expansion of our prior work to investigate tactile exploration policy learning approaches of Robotic palpation for surgical scene localization, 3D surface reconstruction and graphical rendering of sub-dermal tumor geometry augmenting computer assisted surgery. 
The main contributions for developing such tactile navigation policies include: 
\begin{itemize}
    \item A surgical scene initialization and registration pipeline comprising 3D point cloud generated surface mesh  and its corresponding interpolated surface grid mapping.
    \item Bayesian Optimization based palpation coordinate search on the interpolated surface grid locating optimal indentation points.   
    \item Tumor localization policy using subdermal tumor surface contour-following controller governed by impedance control for palpation trajectory generation. 
\end{itemize}
The waypoints from these palpation trajectories are further used for tumor surface 3D reconstructions.   

\begin{figure}[!t] 
\centering
\includegraphics[width= 3.5in]{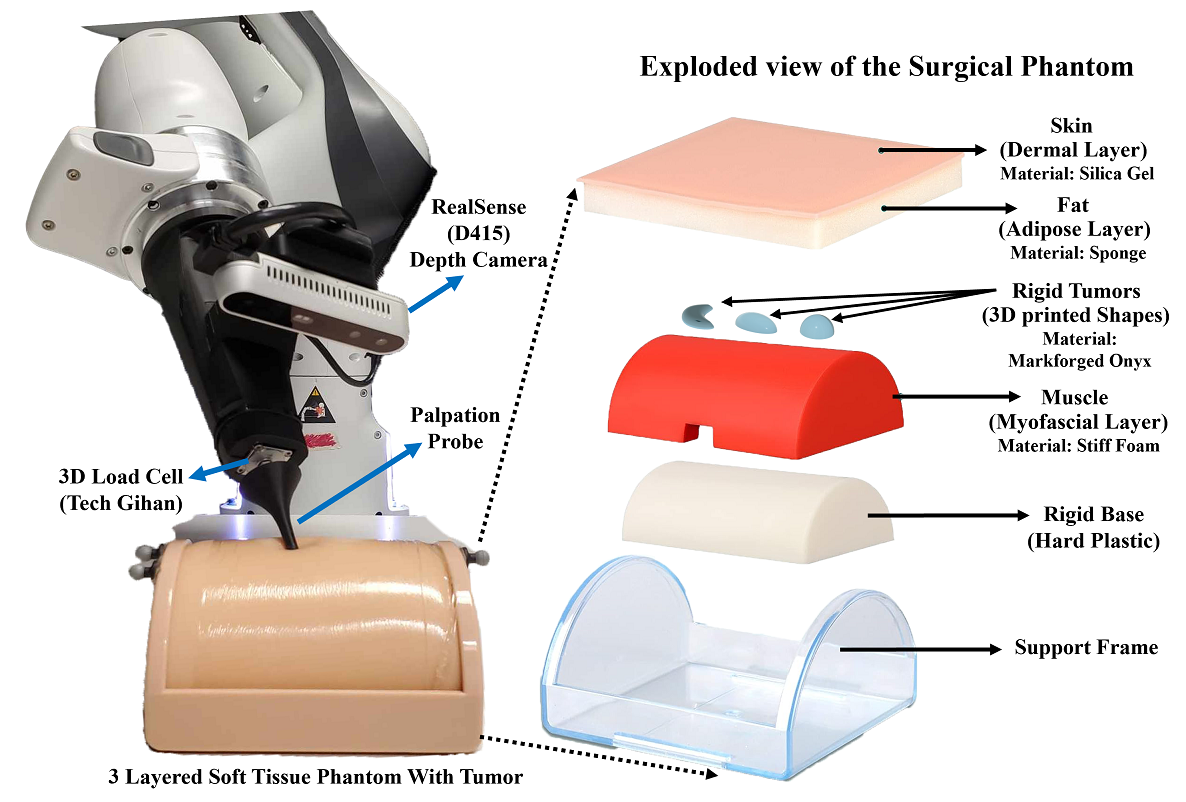}
\caption{The experimental setup of the Robot (7DOF Franka-Panda) with custom built end effector housing a 3D load cell, an IMU sensor and a palpation tip probing a surgical tissue phantom. The phantom is further illustrated with an exploded view which incorporates - the Skin (Dermal Layer) 4mm thickness developed from silica gel, Fat (Adipose Layer) 15mm formed by a sponge material, and muscle (Myofascial Layer in Red) of stiffer foam material covering a rigid base, assembled to a support frame. Tumor geometries of different shapes, 3D printed with Markforged Onyx material, are placed in between muscle and fat layers to experiment 3D reconstruction of the tumor surface profile based on robotic palpation.}
\vspace{-10pt} 
\label{Figure.1}
\end{figure}

\section{System Setup}
Anatomical tissue layers are viscoelastic in nature. Therefore, contact-based robotic tactile exploration engages examining a series of contact parameters (compressive and shear forces, angle of force application, deformation of the tissue and tooltip-tissue interacting friction) that  are necessary to investigate for tissue exploration. 
To simplify our design of experiments, we embed 3D printed rigid tumors of different geometries into tissue layers of a surgical training phantom. 
A 7-DOF Franka-emika Panda collaborative robot arm with a custom designed end effector palpation probe, equipped with custom built force-motion-vision sensing was utilized in carrying out palpation experiments on the phantom (Mediarchitect) \cite{medarchitect}. 
The end effector attachment was packaged with a tri-axial Load-cell (TechGihan Co. Ltd) and a 9 DOF Magnetic, Angular Rate and Gravity (MARG) inertial sensing unit (ICM20948, Invensense) to facilitate online force-motion sensing powered by a 32-bit NXP micro-controller. 
An externally mounted Depth Camera (Intel RealSense D415) served the eyes of the Robot Arm for mapping the external tissue surface point clouds of the phantom. 
The palpation probe includes a funnel-shaped tip with a smooth spherically contoured tissue interaction edge of 5mm diameter similar to RMIS probes for in-vivo tissue explorations. 
The surgical training phantom as shown in Fig.\ref{Figure.1} has a detachable superficial anatomy (Dermal and Adipose layer) with a fixed myofascial (Muscle) layer sitting on a rigid base encapsulated by a support frame. 
Tumors being stiffer than healthy tissues allowed us to 3D print three rigid tumor geometries (Hemisphere, Ellipsoid and Crescent) and screw them into the myofascial layer from the bottom while covering them by the detachable superficial tissue layers to complete the phantom assembly.  
Robot control policies were generated to navigate the visible curved surface contour of the dermal layer to localize and map the underlying invisible subdermal tumor geometries for 3D surface reconstruction and graphical rendering of their shapes using tactile explorations. 

\begin{figure*}[!tbp] 
\centering
\includegraphics[width= 5.5in]{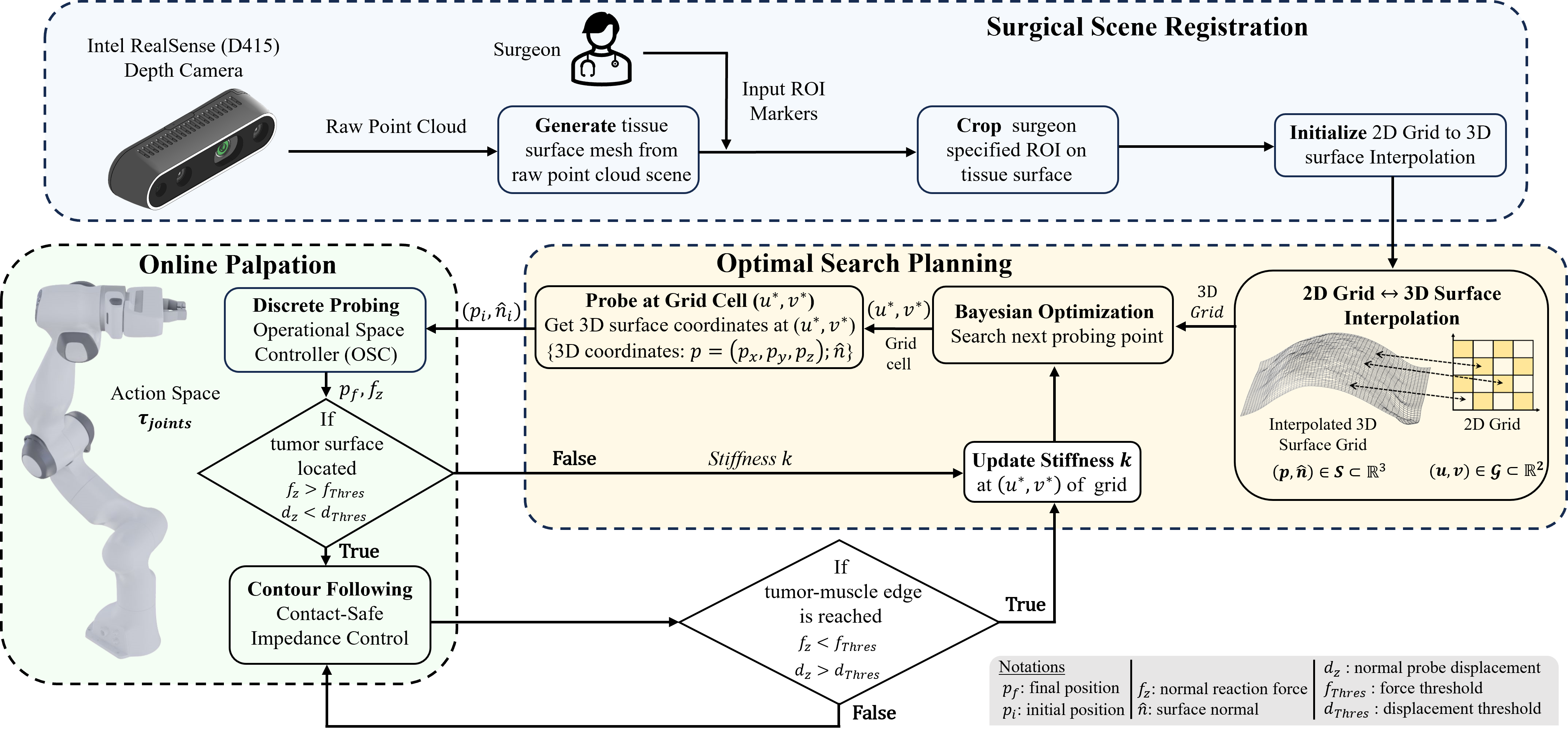}
\caption{ The tactile exploration pipeline for tumor stiffness identification and generation of palpation trajectories. It comprises (a) Surgical Scene Registration by transforming the 3D surface meshes from depth camera point cloud into an interpolated 3D Surface Grid; (b) Optimal Search Planning to locate the next palpation points governed by Bayesian Optimization; (c) Finally the Online Palpation occurs at the searched palpation points from the skin surface to the tumor-muscle boundary using stiffness-estimates based contour following along the tumor contour, which generates the 3D palpation trajectories.}
\vspace{-10pt} 
\label{Figure 2}
\end{figure*}

\section{Methodology}
The automatic tactile exploration policy control pipeline for generating contour following palpation trajectories are shown in Fig.\ref{Figure 2}. This pipeline is further decomposed into surgical scene registration, optimal search planning, and online palpation with gravity compensated force calibration as illustrated below:

\subsection{Surgical Scene Registration}
\begin{figure}[!t]
  \centering
  \vspace{-5pt} 
  \includegraphics[width=0.49\textwidth]{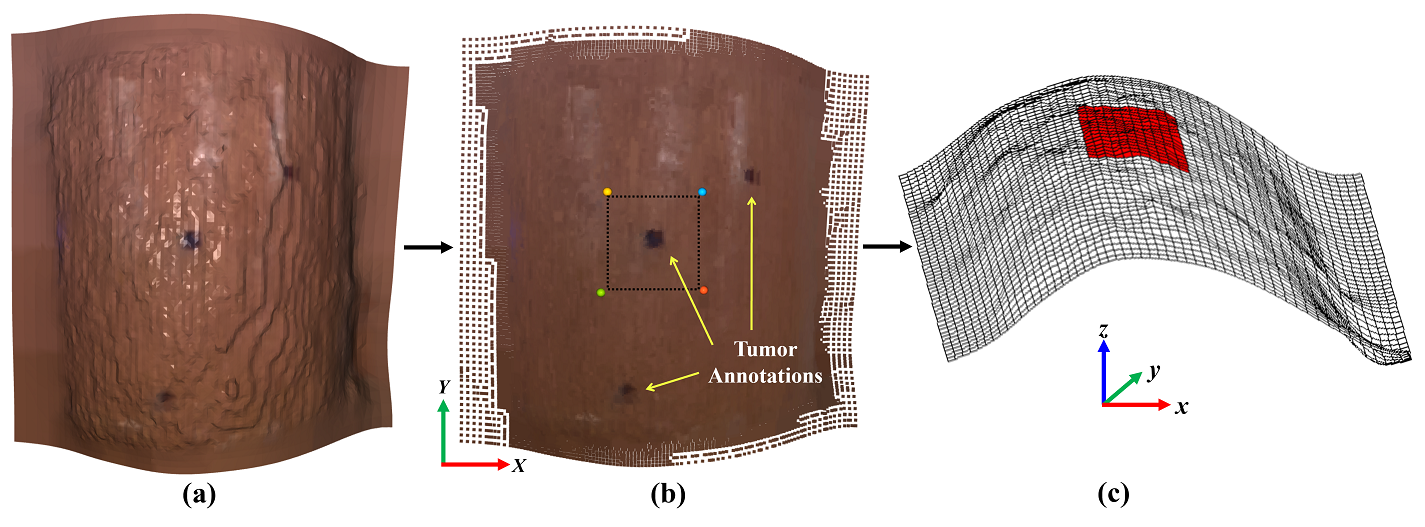}
  \caption{Generation of interpolated 3D surface grid map from surgical scene.  (a) 3D surface mesh created from depth camera's raw point cloud. (b) ROI markers specified by surgeon to create target area bounding box, whereas the dark shaded regions indicate the annotated tumor positions. (c) Generation of Interpolated 3D surface grid with cropped selected ROI (Red) overlayed on the grid cells enclosed by the bounding box.}
  \vspace{-10pt} 
  \label{Figure 3}
\end{figure}

In RMIS, a surgical scene acquired by the depth camera can often be complicated and can only be correctly understood by the surgeon. 
Therefore, our approach offers demarcating a target region of interest (ROI) using a bounding box, by the user (surgeon), reducing the search space. 
This allows the robot to optimally locate tumor coordinates for  successive indentation based palpations. 
For registering the surgical scene, the raw surface point-cloud from the depth camera is post-processed, converted to a smooth 3D triangular surface mesh of the tissue phantom, and finally cropped into a user specified ROI. This ROI of the phantom's surface mesh is enclosed in a bounding box for further interpolation into a 3D surface grid map as shown in Fig. \ref{Figure 3} . 


Given the objective to minimize palpations, discrete search in Cartesian Space can be computationally expensive.
Additionally, the points from the 3D surface mesh are unstructured, nonuniform and scattered, leading to difficulty in sampling for search space optimization. Therefore a uniform 2D grid with sampling resolution $(dx, dy)$ is generated within the external bounds of X-Y plane (Fig. \ref{Figure 3}(b)) of the surface mesh. This 2D grid is interpolated to the scattered points in the 3D surface mesh using the cubic interpolation method generating a 3D surface grid map. Vertices of each grid cell $(u,v) \in \mathcal{G}$ corresponds to a point and normal $(p,\hat{n})$ on 3D surface mesh $\mathcal{S}$.
The grid cells of this interpolated surface grid is further used in optimal search planning to determine palpation coordinates.

\begin{figure*}
\centering
{\includegraphics[width=0.33\textwidth]{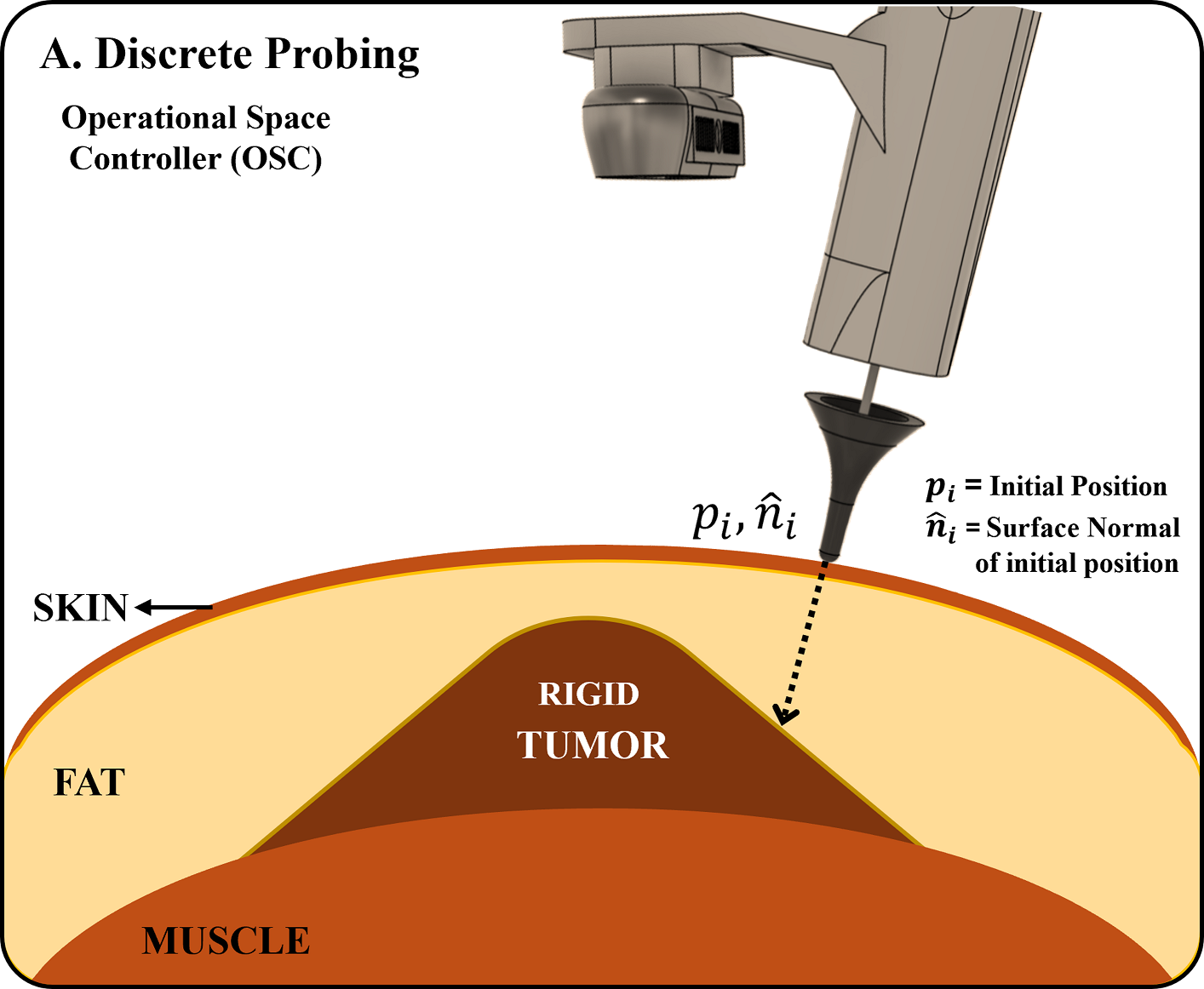}}%
\hfill 
{\includegraphics[width=0.33\textwidth]{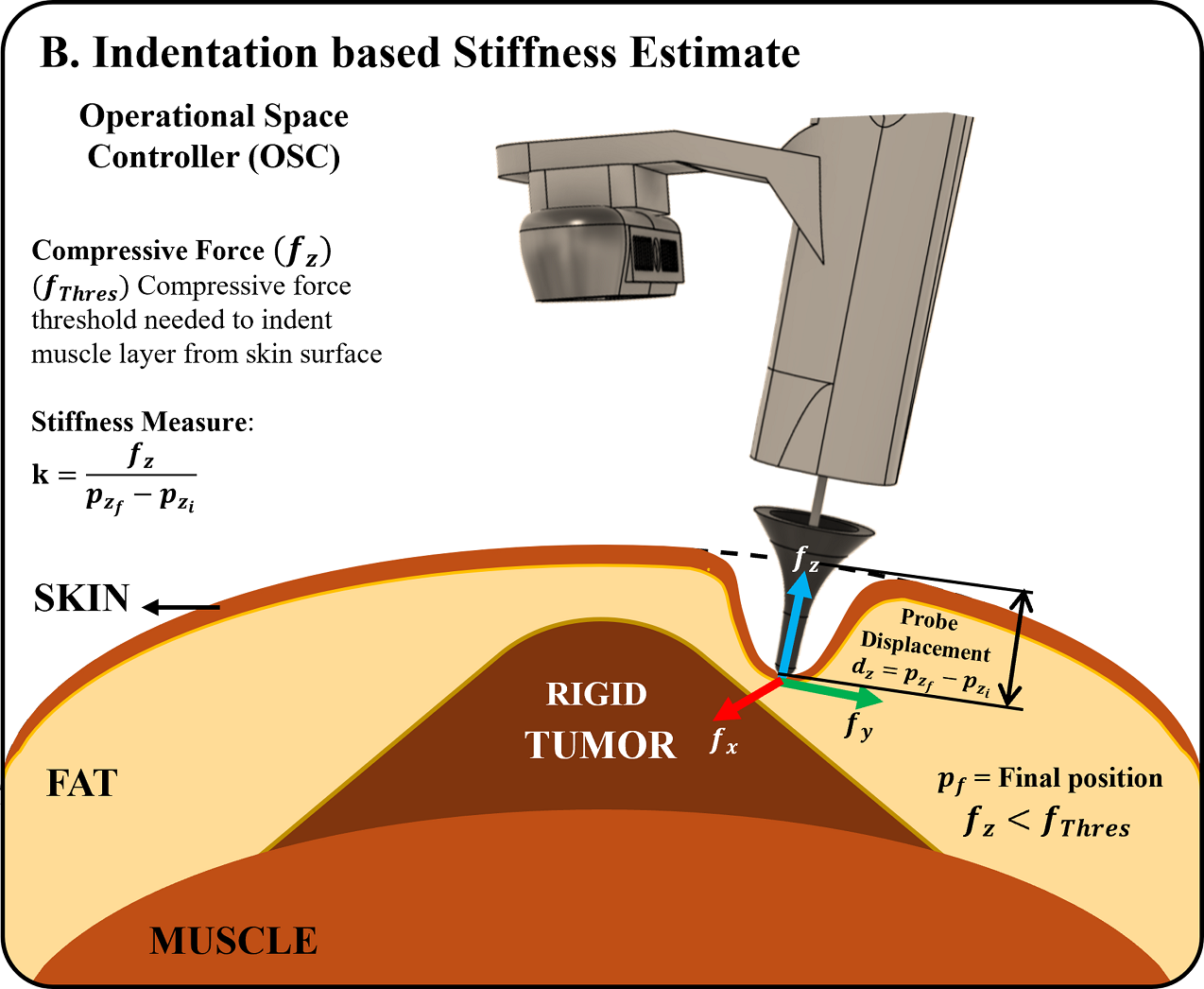}}%
\hfill 
{\includegraphics[width=0.33\textwidth]{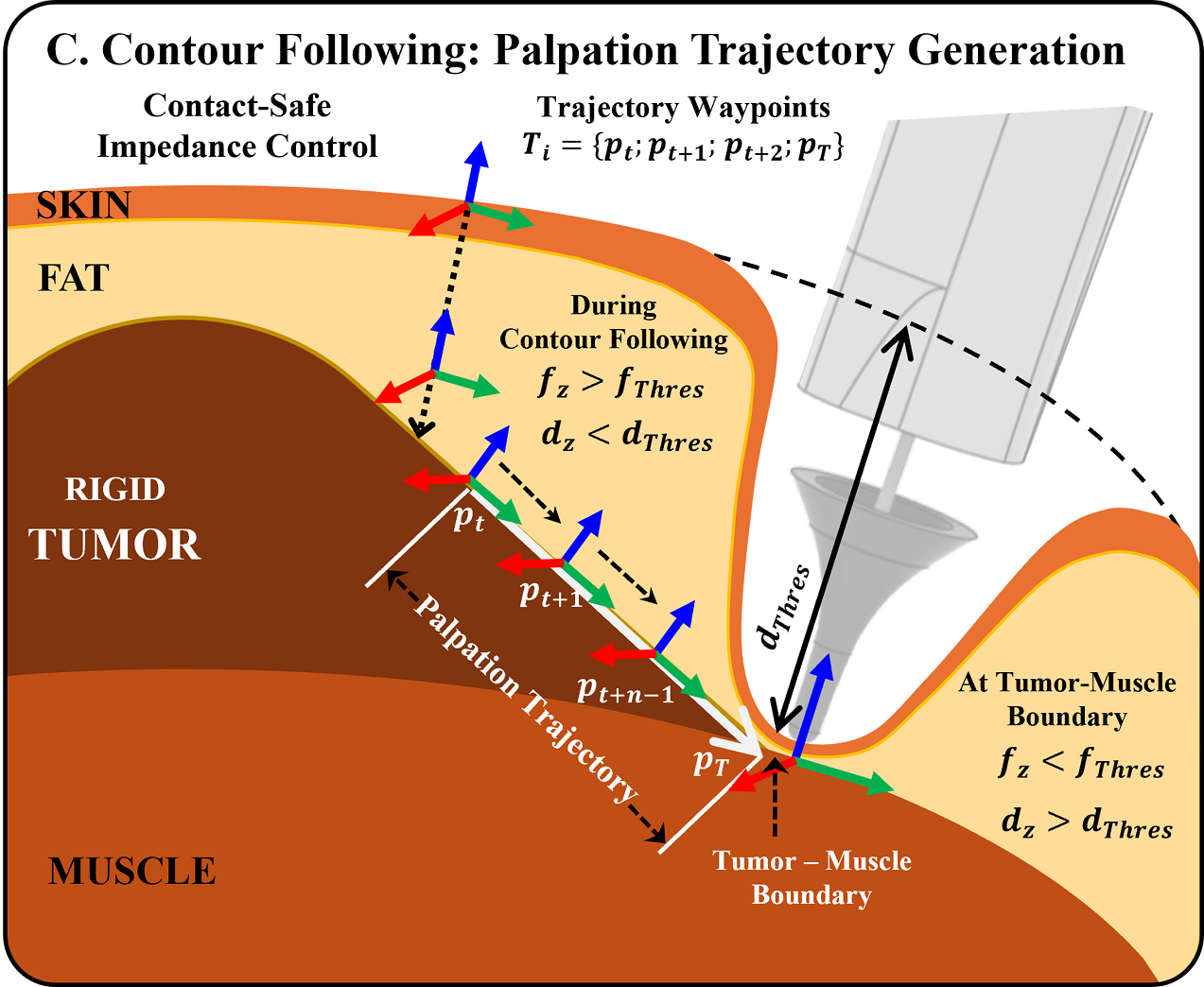}}%
\caption{Representation of online palpation to detect presence of tumor using probing and palpation trajectory generation using contour following. (A) For Probing, Operational Space controller (OSC) is commanded to orient along the surface normal and indent at the optimal grid cell to detect presence of tumor. (B) The stiffness estimate is generated at the probed coordinate. If the force at maximum probe displacement exceeds a force threshold then tumor presence is confirmed at that coordinate. (C) If tumor presence is detected, then controller switches to contact-safe impedance control for tumor contour following. The probe travels along the contour of the tumor following the force threshold condition to reach the tumor-muscle boundary and terminates contour following. The trajectory way-points of the contour following yields the palpation trajectory further used for tumor surface 3D reconstruction.}
\vspace{-10pt} 
\label{Figure 4}
\end{figure*}

\subsection{Optimal Search Planning}
Random search of palpation points may affect the robustness of exploring the tissue surface leading to discretely probing the phantom without locating the tumor in minimal time. 
Therefore, optimal search planning was necessary to locate coordinates  with higher stiffness values representing a tumor within a given probe displacement range.  
To achieve this we adopted searching palpation coordinates on the surface 2D grid with Gaussian Process regression and Bayesian Optimization.

\subsubsection{Gaussian Process Regression}
Gaussian processes (GP) are commonly used to model the stiffness distributions expanded over the total sample space of probed cells on the surface grid. 
The stiffness at each probed cell $(u,v) \in \mathcal{G}$ on the grid is represented as a random variable sampled from a gaussian distribution $N(\mu(k),\sigma)$, where $\mu(k)$ is the mean and $\sigma$ is the variance involved, and $k$ is the stiffness measures observed as a random variable for successive indentations. GP regression, adopted from \cite{salman2018trajectory},  makes predictions for next probed cell $(u^*,v^*)$ for maximum observed stiffness. 

\subsubsection{Bayesian Optimization}
Bayesian optimization (BO) allows extracting the global maxima of a distribution function from its samples. 
Therefore probing at different coordinates on the surface grid produces a stiffness distribution sample space with stiffness $k$ for at least two or more indentations.
BO selects the optimal stiffness state $k^*$, using the posterior mean $\mu(k)$ of the GP, which corresponds to grid cell $(u^*,v^*)$ that optimizes over an Expected Improvement (EI) objective function. For further details, mathematical expressions of this objective function are well elaborated in \cite{salman2018trajectory}. 
This is how the tumor coordinates are detected in the given search space, which allows BO to collect more samples around its vicinity to facilitate dense palpation for refined tumor surface reconstruction.

\subsection{Force Calibration and Gravity Compensation}
Gravity compensation is essential for calibration and accurate force estimation at the tip of the palpation probe. 
As force sensing from the Load Cell as well as force estimation from the Robot's joint torques were utilized for tactile sensing, we included a two-fold gravity compensation method. The moment of inertia matrix $I_{eef}$ of the robots end effector attachment, including its centre of mass were acquired from CAD. The mass of the attachment were determined from weighing the full assembly of the attachment. These parameters were fed into the Franka Panda's Low level gravity compensation and state estimation algorithms to facilitate accurate force estimation at the tip from joint torques. 
Similarly gravity compensation was conducted for force calibration of the triaxial load cell eliminating force skewness due to tension exerted by the palpation tip. 
This is performed in two steps – (a) Load cell Z axis voltage offset shift compensation due to tension forces exerted by palpation tip attachment; and (b) Weight compensation of the tip using forward-inverse kinematics of the Load cell coordinate frames in three steps. 
The weight compensation is elaborated as:\textit{(i)} Forward kinematics performed by transformation of the Load cell co-ordinates from Local to Inertial reference: $\overrightarrow {\textbf{\textit{f}}_{I}^{L}} = [\textbf{R}_{\psi,\theta^{'},\phi}] \times \left( \overrightarrow {\textit{f}_{(x,y,z)}} \right)$, where $[\textbf{R}_{\psi,\theta^{'},\phi}]$ is the rotation matrix of $\mathbb{R} \in \textit{SE(3)}$ euclidean space in ZYX Euler combination and $\psi,\theta,\phi$ are Euler angles (Yaw, Pitch, Roll) computed from sensor fusion of accelerometer and gyroscope readings of the IMU sensor; \textit{(ii)} Subtraction of the tip mass $M_{L}$ Newtons from the load-cell Z direction in the inertial frame by: $\overrightarrow {\textbf{\textit{\textit{f}}}_{I}^{L}} = \overrightarrow {\textbf{\textit{f}}_{I}^{L}} - \begin{bmatrix} 0 & 0 & M_{L}\end{bmatrix}^{T}$; and \textit{(iii)} Finally inverse kinematics is used for transforming the Load-cell updated force values back to the local reference from the inertial reference by: $\overrightarrow {\textbf{\textit{\textit{f}}}_{L}^{I}} = [\textbf{R}_{\psi,\theta^{'},\phi}]^{-1} \times \overrightarrow {\textbf{\textit{f}}_{I}^{L}} $. The robustness of this force calibration with gravity correction approach has been validated in handheld devices \cite{bhattacharjee2022handheld} of manual therapy. The Root mean squared value computed from $\overrightarrow {\textbf{\textit{f}}_{L}^{I}}$ serves as the load cell derived resultant reaction force  acting at the tip of the palpation probe used in further computations.

\subsection{Online Palpation}
Our online palpation framework has two primitives: Discrete Probing to detect tumor coordinates on the Surface grid and Contour following along the tumor surface geometry to generate palpation trajectories. 


\subsubsection{Discrete Probing}
Operational Space Controller (OSC) \cite{khatib1978} facilitates safe generation of command joint torques $\tau_{joint}$ providing a sufficient tradeoff between passivity and motion accuracy for contact-rich applications \cite{nakanishi2008operational}.
Therefore, OSC is utilized for moving and orienting the end-effector tip at the initial probing point on the surface grid cell for discrete probing based indentations.
The axial direction of the tip is aligned along the surface normal $\hat{n}_i$ of initial probing point $p_i$ acquired from the grid to surface map, as shown in Fig. \ref{Figure 4} (a).  
This is followed by compressive indentation of the tissue along the surface normal directions to generate stiffness estimates using compressive reaction force component $f_{z}$, and probe displacement $d_{z}$ shown in Fig. \ref{Figure 4} (b), along the z axis of the load cell. 

\subsubsection{Stiffness Estimate}
The external load cell derived forces as well as end-effector forces computed by the Robot's joint torques were leveraged to compute stiffness estimates of each probed coordinate on the surface grid cell. The gravity corrected calibrated force vector was primarily utilised further in the development pipeline. Compressive force threshold $f_{Thres}$ across the surface indentation depth $d_{Thres}$ from skin to muscle layer was experimentally computed. The end-effector indentation will either approximate displacement from skin to tumor or displacement to muscle at the termination point $p_{z_f}$ from initial point $p_{z_i}$ for each probing coordinate's $p_z$ component. 
The end-effector indentation depths are calculated from probe displacements $d_z = p_{z_f}-p_{z_i}$ opposite to the surface normals applying compressive reaction forces $f_{z}(t)$, where compressive force $f_{z}$ is a function of time $t$.
Stiffness estimates at each indentation is computed as: $k = \frac{f_{z}}{d_z}$, which determines whether the indentation touches a tumor surface or a muscle layer.
If the condition for indentation force $f_{z}(t) > f_{Thres}$  and probe displacement $d_z < d_{Thres}$ is met, then the indentation confirms detection of a tumor coordinate, which needs to be succeeded by the contour following primitive. 

\subsubsection{Contour Following}
Once a tumor coordinate is detected as a result of a discretely probed indentation, and the tip is in contact with the skin, the OSC switches to contact safe impedance controller.
This impedance controller generates motions to trace sub-dermal surface profile of the tumor geometry until it reaches the tumor-muscle boundary for palpation trajectory generation.

To generate contour following controller targets, we empirically derived a
simple, oscillatory motion path $\Delta p_{xy}$ with minimal jerk for tracing the tumor surface contour until it reaches the tumor-muscle boundary in iterative time-steps, without getting stuck at each iteration. 
\begin{align}
\Delta p_{xy}(t) &= 2A \cdot (10t^3 - 15t^4 + 6t^5) - A
\end{align}
where $\Delta p_{xy}(t)$ is a small finite pose change due to oscillatory motion to trace the tumor surface along $x$ (red) or $y$ (green) direction of the contour (Fig. \ref{Figure 4}(c)) while maintaining compressive indentation and contact with tissue surface. Additionally, $A$ is the amplitude of the oscillatory motion and $t$ is the time-step sampled at a frequency of $80Hz$. At every iteration the upstream controller targets a desired position $\mathbf{p_d}$, from current position $p(t)$  where:
\begin{align}
\mathbf{p_d} &= p(t) + \Delta {p_{xy}(t)}
\end{align}

Contour following requires safe, contact-rich interactions with the tissue, while avoiding potential unsafe accelerations. The standard impedance control objective \cite{impedanceNevile} explicitly modulates a commanded end-effector force $\mathbf{f_{cmd}}$ to model contact rich interactions.  
\begin{align}
\mathbf{f_{cmd}} &= K_d \cdot (\mathbf{\dot{p}_d} - \mathbf{\dot{p}}) + K_p \cdot (\mathbf{p_{d}} - \mathbf{p})
\end{align}
where $\mathbf{f_{cmd}}$ is the commanded end-effector force opposite to $f_z$ direction (Fig. \ref{Figure 4} (b)), following the principle of dynamic equation of motions by modelling a spring-damper equilibrium system with $K_d$ and $K_p$ being the damping coefficient and spring constant respectively. Here, $\mathbf{p_{d}}$ is the desired pose and $\mathbf{p}$ is the measured pose of the end-effector with pose error $ \mathcal{E} = \mathbf{p_{d}} - \mathbf{p}$ and velocity/twist error $ \dot{\mathcal{E}} = \dot{\mathbf{p_{d}}} - \dot{\mathbf{p}}$ . This commanded interaction force objective is further transformed into joint torques $\mathbf{\tau}_{joints}$
\begin{figure}[!t]
  \centering
  \includegraphics[width=0.48\textwidth]{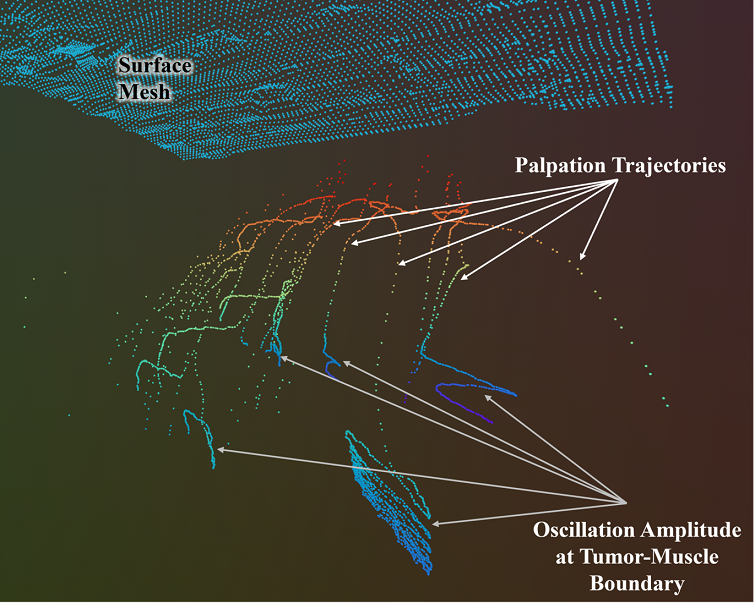}
  \caption{3D visualization of generated palpation trajectories from contour-following of the tumor surface, which takes the shape of the tumor's surface profile. The oscillations with higher amplitude is observed at the edge of the tumor i.e. tumor-muscle boundary. The cropped surface mesh of the phantom is displayed above the tumor surface}
  \vspace{-10pt} 
  \label{Figure 5}
\end{figure}
\begin{align}
\mathbf{\tau}_{joints} &= J^T \mathbf{f_{cmd}} + \mathbf{\tau}_{corr}
\end{align}
where $J$ is the robot Jacobian matrix, 
$\mathbf{\tau}_{corr}$ is the coriolis joint torques, while 
a bounded pose error limit $ \mathcal{E}_{Thres}$ is asserted to ensure safe interaction of the tissue-tip contact during contour following.
This limits the commanded interaction forces withing an admissible range without damaging the tissue while maintaining optimal contact as described in \cite{luo2024serl}. The admissible interaction force $\mathbf{f_{adm}} = K_p \cdot |\mathcal{E}_{Thres}| + \frac{2 \cdot K_d \cdot |\mathcal{E}_{Thres}|}{T}$, where $T$ is the controller period $(1ms$).

\begin{figure*}
    \centering
   \includegraphics[width=1.0\textwidth]{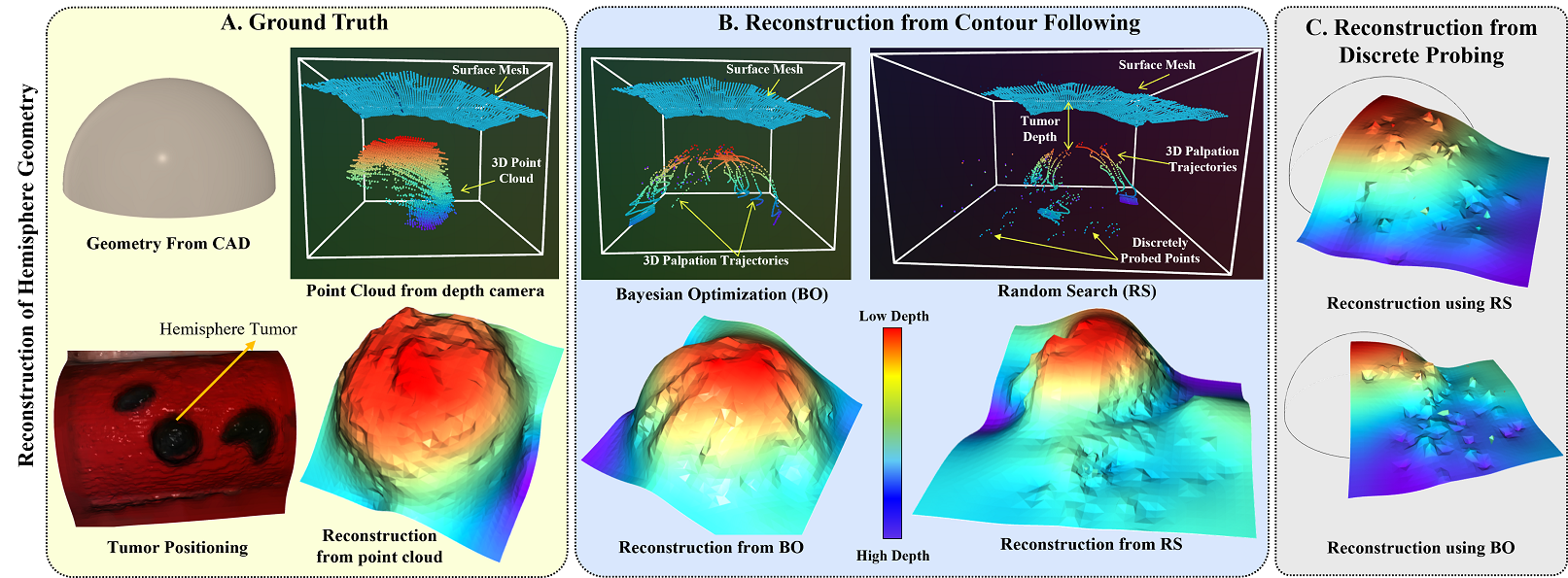}
   \\
   \includegraphics[width=1.0\textwidth]{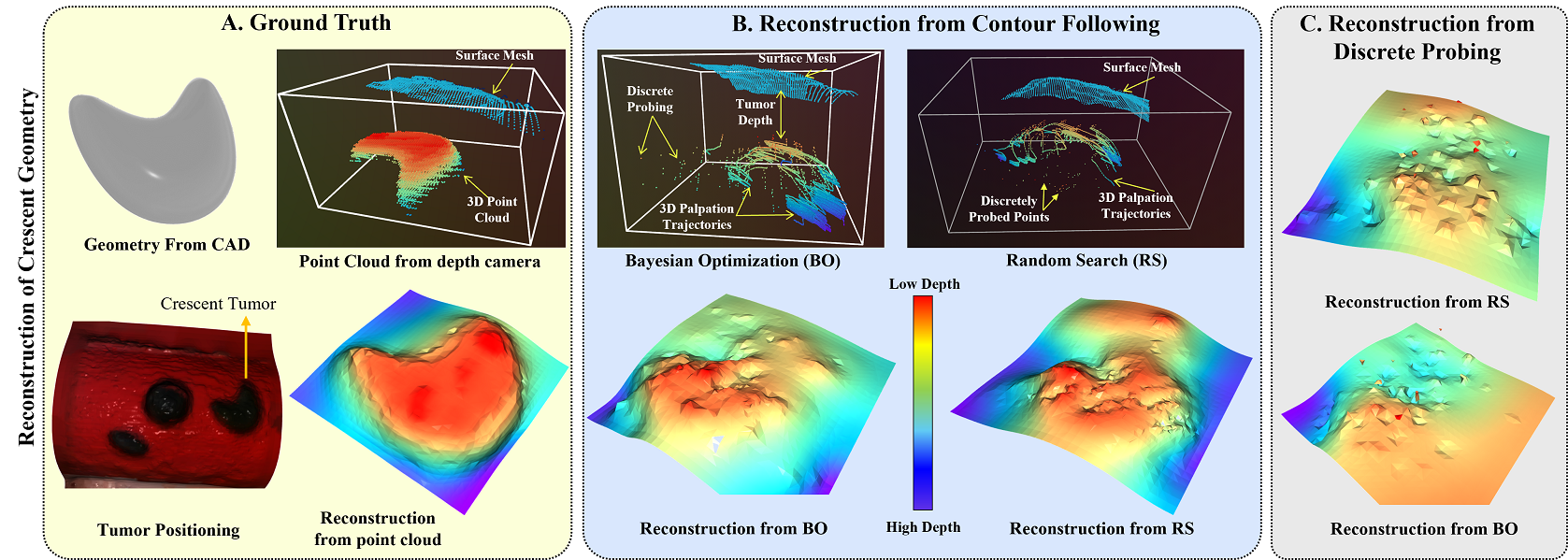}   
    \caption{Reconstruction of tumor geometries using different palpation approaches (Contour following and Discrete probing). (A) Ground truth representation of the hemisphere and crescent geometry. The positioning of tumor geometries (black) placed on the muscle layer (Red) of the phantom is shown. Surface mesh of tumor positioning is acquired from the depth camera renders by separating the skin-fat layer attachment for extracting the ground truth. 3D plots of tumor geometry point clouds represent the ground truth. (B) 3D reconstruction of the tumor surface geometries using contour following. The palpation trajectories governed by the Bayesian Optimization (left column) approximates a complete reconstruction compared to that of Random Search (Right column) without BO. (C) Representation of reconstruction results from Discrete Probing without contour following where, sampling with Random Search (Top) shows a better geometry compared to probing by Bayesian Optimization (bottom), for Crescent geometry, while partial reconstructions are represented for the hemisphere geometry based on limited palpations. Visualizations made with \cite{RerunSDK} \cite{Zhou2018}.}
  \vspace{-10pt} 
    \label{Fig.6}
\end{figure*}

 The contour following impedance controller terminates when the indentation reaches the tumor-muscle boundary and meets the condition of indentation depth $d_z > d_{Thres}$ and indentation force $f_{z}(t) < f_{Thres}$ as it indents the compliant muscle layer as shown in Fig. \ref{Figure 4}(c).

\subsection{Palpation Trajectories}
 
 The pose $p(t)$ extracted at every iteration time step are 3D plotted to generate the palpation trajectories of the contour following oscillatory motions. At the tumor-muscle boundary the compressive force condition switches from $f_{z}(t) > f_{Thres}$ to $f_{z}(t) < f_{Thres}$, due to the compliance of the conforming muscle layer. Therefore a high amplitude of oscillatory motions are observed at the end of each palpation trajectory as shown in Fig. \ref{Figure 5}. Additionally, the pose error $\mathcal{E}$ drastically increases exceeding the error limit and forcing the impedance controller to fail. An exception handling method resolves this limitation by adding timeout function which automatically terminates the contour following and transfers the control to OSC for probing next palpation points. The palpation trajectories generated from each contour following motions are projected into a 3D triangular surface mesh to 3D reconstruct the tumor geometry surface plots.

\section{Experiments and Results}
Two different palpation approaches were adopted to test surface 3D reconstructions of tumor geometries. The first approach was discretely probed indentations of the cropped phantom surface with Bayesian Optimization (BO) or Random search (RS) without BO, both of  which doesn't involve palpation trajectories. While, the second approach involves BO or RS based probing for search space sampling to detect high stiffness points and then continuously tracing the stiffer geometry using contour following indentation path to generate palpation trajectories. For experimental convenience the embedded tumors were screwed into the muscle layer from the bottom, while the phantom assembly after integrating the top tissue layers were fixed to the base frame with an adhesive. This ensured eliminating all internal and external motions of the phantom and tumours during palpation. We performed several experimental palpation trials on cropped ROIs of the annotated hemisphere and crescent tumor geometry respectively. The ellipsoid geometry showed similar reconstructions compared to the hemisphere geometry without significant distinctions hence we discarded it from our trials. Some of the experimental assumptions and limitations are enlisted below:
\begin{itemize}
    \item The tri-layered phantom with tumor embeddings are assumed to have stiffnesses of the form: $(k_{Fat} < k_{Skin} < k_{muscle} < k_{tumor})$. The phantom surface and tissue layers are conformal with large deformations. 
    \item  Friction between tool-tip and phantom surface is not considered, whereas friction between tissue layers are also neglected. For contact safe impedance control and the robot's convenience we applied medical grade cellophane wrap on the phantom surface to prevent the skin from ripping apart and damaging the tissue layers.  
    \item An actual RMIS surgical scene is kinematically constrained, limiting the workspace of the robot. For simplicity, this study does not consider these additional workspace constraints during selection of palpation poses on the tumor surface and during contour following.
\end{itemize}

The evaluation of the surface reconstructions as compared to ground truth, which is further illustrated below. 

\subsection{Evaluation Metrics}
For comparing the reconstruction with the ground truth we used an F-Score metric, as established in geometry reconstruction works \cite{suresh2023neuralfeels, knapitsch2017tanks, tatarchenko2019single}.
Precision is a similarity metric of the reconstructed points compared to the points in the ground truth using a distance threshold $r$. 
Whereas Recall metric evaluates the similarity percentage of the ground truth points compared to the reconstructed points using $r$, thereby eliminating false positives. A Harmonic mean of the recall and precision measures finally constitute the F-Score metric, capturing both the surface reconstruction accuracy and shape completion using an interpretable value from $0-1$, where $0$ represents no similarity and $1$ indicates complete similarity. 

\begin{table}[ht]
    \centering
    \caption{Tumor 3D Reconstruction Error}
    \begin{threeparttable}
        \begin{tabular}{lrrrrrr}
            \toprule
            & \multicolumn{3}{c}{Hemisphere} & \multicolumn{3}{c}{Crescent} \\
            \cmidrule(lr){2-4} \cmidrule(lr){5-7}
            & Mean & Max &\#P & Mean & Max & \#P\\
            \midrule
            RS, with CF (Ours) & \textbf{0.61} & 0.81 & 50 & \textbf{0.94} & 0.95 & 80 \\
            RS, w/o CF & 0.48 & 0.55 & 50 & \textbf{0.94} & 0.95 & 80 \\
            BO with CF (Ours) & 0.57 & \textbf{0.89} & 50 & 0.89 & 0.94 & 80 \\
            BO w/o CF  & 0.46 & 0.62 & 50 & 0.90 & \textit{\textbf{0.96}} & 80 \\
            \bottomrule
        \end{tabular}
        \begin{tablenotes}
            \small
            \item Note: RS = Random Search, CF = Contour Following, w/o CF = Discrete Probing, \#P = Number of palpations, Mean = Average F-Scores computed with \(r=3mm\) and with 10 trials for each tumor shape, Max = Maximum computed F-Score taken from the 10 trials.
        \end{tablenotes}
    \end{threeparttable}
     \vspace{-15pt} 
    \label{Table.1}
\end{table}

\subsection{Tumor Surface Reconstructions}
 Both the hemisphere and crescent ($radius = 1cm$) tumor geometries were palpated for around 10 trials each on BO and RS based sampling for both discrete probing and contour following indentation approaches evident from Table \ref{Table.1}, at an admissible force range $(5\sim10N)$. The series of reconstructions with their corresponding trajectory plots and reconstructed meshes are shown in Fig. \ref{Fig.6}. From visual observations, the palpation with contour following indentations sampled by BO produces the best results for hemisphere geometry complimented by the maximum F-Score of $89\%$ precision. However the reconstruction performance reduces as we move from contour following to discrete probing. This is because for $n$ palpations using discrete probing we get only $n$ stiffness points for reconstruction which gets increased by $10X$ while using contour following, as they generate $10-20$ trajectory way-points for each palpation. Therefore Discrete probing will need $>100$ palpations to better reconstruct underlying stiffer subsurface geometry which can be halved by contour following indentations. However, the geometry of the Crescent shape is mostly flat except at the edges. Hence reconstructions with RS based sampling produces similar results with discrete probing or contour following. Nonetheless, BO sampled reconstructions with discrete probing produce continuous planes instead of a contour which matches the top flat surface in ground-truth mesh resulting in high F-Scores $(0.96)$. The combined F-Scores taking palpation points from all performed trials for Hemisphere geometry ($\#P=2000$) and Crescent geometry ($\#P=3200$) are 0.72 and 0.92, respectively, which can benchmark reconstructions from our trials for $<100$ palpations. Hence, modifying distance parameter $r$ and increasing palpations $\#P$ can improve F-Score performance. 

\section{Discussions and Conclusion}
This study produces a novel tactile exploration policy for surface 3D reconstructions of embedded subdermal tumor geometries in a tri-layered curved surface tissue phantom using robotic palpation aimed towards RMIS applications.
The tumor surface contour following continuous indentation approach yields the palpation trajectories by tracing the tumor surface geometry along continuous palpation path reaching the tumor-muscle boundary in $< 100$ palpations. 
3D plots of these trajectory way-points facilitates generating the reconstructed 3D surface mesh of targeted tumor geometry, which accounts to the primary contribution of this study.

The experimental results and visual observations from several palpation trials shows that tumor geometry contour following indentation approach outperforms the discrete probing approach in reconstructing the surface profile. 
Contour following offers a combination of two type of indentations - discrete probing (skin and fat layers) to locate tumor coordinates and continuous indentations based tumor geometry tracing to retrieve the shape of the tumor. 
Theoretically, BO based search space sampling to initiate contour following should produce better reconstructions as compared to RS based outcomes.
However, discontinuity in the reconstructed geometries are observed from both sampling approaches for the contour following indentation approach. 
This may be due to the drifts or sampling errors in the surface normal alignments of the end effector  which resulted in sampling stiffer points in close proximity to last sampled point instead of covering the total cropped ROI. 

However, improvements can be made to the current pipeline by introducing reinforcement learning based objective functions to determine the north and east direction of the probed points and un-probed locations. 
This approach might allow some uniformity in the sampling across the area in stipulated number of palpations to generate 3D reconstruction of a continuous surface geometry minimizing discontinuities. 
This work does not model friction directly, which is essential for following Cartesian paths maintaining optimal contact-safe indentations over the soft tissue surface. 
In RMIS, in vivo intraoperative surgical scenes involve environment constraints, which restricts kinematic motions in the feasible configuration space of the robot manipulator, as well as limits search space on the soft tissue. 

In future works, this can be addressed by adding state-of-the-art confined-space motion planning algorithms to the current pipeline or transferring the pipeline to other continuum robot manipulators with a better designed kinematic workspace for surgical tasks. This study was conducted on rigid tumor inclusions which will be extended to semi-rigid and soft tumors with more complex geometries and variety of sizes in the future for bench-marking the robustness of the tactile exploration policy and reconstruction pipeline. 

Moreover, this tactile exploration policy for tumor 3D reconstruction can be translated to clinically applied robotic laparoscopic surgeries on existing surgical robot setup e.g. multiport da Vinci XI from Intuitive Surgical or single port surgical robot from Vicarious surgical and others, by integrating a miniaturised tri-axial force sensor at one of the probe tips. The on-arm stereo camera for single port robot or the endoscope attached to a multiport da Vinci can capture point clouds of the surface topology of intraoperative space while the tactile exploration pipeline can be deployed to map and reconstruct subsurface tumor geometries aimed for future clinical translations.  

\section{Acknowledgements}

We thank Zhengtong Xu and Xinwei Zhang for helpful technical discussion and feedback throughout the project. We thank Sheeraz Athar for feedback on the palpation probe design. This work was partially supported by Showalter Trust. 

\bibliographystyle{IEEEtran}
\bibliography{bibtex/bib/IEEEabrv,paper}
\end{document}